\documentclass[conference]{IEEEtran}
\IEEEoverridecommandlockouts
\usepackage{amsmath,amssymb,amsfonts}
\usepackage{algorithm}
\usepackage{graphicx}
\usepackage{textcomp}
\usepackage{xcolor}
\usepackage{booktabs}
\usepackage{multirow}
\usepackage{subfigure}
\usepackage{pifont}
\usepackage{xspace}
\usepackage{hyperref}
\usepackage{algpseudocode}
\usepackage{color}
\usepackage{bm}
\usepackage{tabularx}
\usepackage{bm}
\usepackage[square,sort,comma,numbers]{natbib}
\usepackage{caption}

\hypersetup{
    colorlinks=true,
}
\def\BibTeX{{\rm B\kern-.05em{\sc i\kern-.025em b}\kern-.08em
    T\kern-.1667em\lower.7ex\hbox{E}\kern-.125emX}}
\begin{document}

\title{
    FPPL: An Efficient and Non-IID Robust\\ Federated Continual Learning Framework\\
    \thanks{This work was supported by the STCSM (22QB1402100) and NSFC (62231019, 12071145). Corresponding author: Xiangfeng Wang.}
}

\author{
    \IEEEauthorblockN{
        Yuchen HE\IEEEauthorrefmark{1},
        Chuyun SHEN\IEEEauthorrefmark{1},
        Xiangfeng WANG\IEEEauthorrefmark{2}\IEEEauthorrefmark{1},
        Bo JIN\IEEEauthorrefmark{3},
    }
    \IEEEauthorblockA{
        \IEEEauthorrefmark{1}School of Computer Science and Technology, East China Normal University, Shanghai, China 200062\\
        \IEEEauthorrefmark{2}Shanghai Formal-Tech Information Technology Co., Lt, Shanghai, China 200062\\
        \IEEEauthorrefmark{3}School of Software Engineering, TongJi University, Shanghai, China 200092
    }
}

\maketitle

\begin{abstract}

    Federated continual learning (FCL) aims to learn from sequential data stream in the decentralized federated learning setting,
    while simultaneously mitigating the catastrophic forgetting issue in classical continual learning.
    Existing FCL methods usually employ typical rehearsal mechanisms,
    which could result in privacy violations or additional onerous storage and computational burdens.
    In this work, an efficient and non-IID robust federated continual learning framework,
    called \textbf{F}ederated \textbf{P}rototype-Augmented \textbf{P}rompt \textbf{L}earning (FPPL), is proposed.
    The FPPL can collaboratively learn lightweight prompts augmented by prototypes without rehearsal.
    On the client side, a fusion function is employed to fully leverage the knowledge contained in task-specific prompts for alleviating catastrophic forgetting.
    Additionally, global prototypes aggregated from the server are used to obtain unified representation through contrastive learning,
    mitigating the impact of non-IID-derived data heterogeneity.
    On the server side, locally uploaded prototypes are utilized to perform debiasing on the classifier,
    further alleviating the performance degradation caused by both non-IID and catastrophic forgetting.
    Empirical evaluations demonstrate the effectiveness of FPPL,
    achieving notable performance with an efficient design while remaining robust to diverse non-IID degrees.
    Code is available at: \url{https://github.com/ycheoo/FPPL}.

\end{abstract}


\section{Introduction}
\label{sec:intro}

Federated Learning (FL)~\cite{mcmahan2017communication,yang2019federated,kairouz2021advances,huang2024federated} is considered as a specific distributed learning paradigm
in which clients update their models by aggregating information from different client-side local models without uploading private data, thereby ensuring data privacy.
Most existing FL methods assume that the data of clients is static without dynamically changing over time.
However, in real-world applications, the data of clients accumulates continuously,
which would lead to the emergence of new classes or potential shifts in data distribution.
Typical static FL methods become not applicable,
the catastrophic forgetting issue, degrading performance on previous tasks,
would significantly occur when data dynamically grows~\cite{kirkpatrick2017overcoming,rebuffi2017icarl},
which could be exacerbated by the non-IID (non-independent and identically distributed) issue of FL~\cite{zhao2018federated,li2022federated}.

Therefore, it becomes natural to consider the Federated Continual Learning (FCL)~\cite{yoon2021federated,dong2022fcil}.
FCL aims to establish federated learning methods that can adapt to new data streams and mitigate catastrophic forgetting~\cite{rebuffi2017icarl} on previous tasks.

Prevalent FCL methods typically rely on rehearsal~\cite{dong2022fcil,qi2023better},
a common technique in continual learning~\cite{rebuffi2017icarl,wu2019bic,zhou2023memo},
where data from prior tasks is utilized for subsequent training to mitigate catastrophic forgetting.
This practice, integrating previous task data into the current task,
raises privacy concerns or adds storage and computational overhead in FCL.
For instance, GLFC~\cite{dong2022fcil} locally store previous task data for rehearsal,
posing privacy risks and introducing additional burdens on clients.
Moreover, FedCIL~\cite{qi2023better} uses generative replay to produce synthetic data,
avoiding raw data maintenance but still incurring additional storage and computational cost due to the use of generative models.
As a consequence, rehearsal-based methods lack scalability when faced with an increasing number of new tasks.
Instead, the most efficient approach is to update the model by training solely on new data,
but such a rehearsal-free method may result in catastrophic forgetting.

Fortunately, leveraging pre-trained transformers~\cite{vaswani2017attention,devlin2019bert,dosovitskiy2021vit} has proven to be a promising approach for mitigating catastrophic forgetting without the need for rehearsal.
For instance, Fed-CPrompt~\cite{bagwe2023fedcprompt} proposes a rehearsal-free FCL approach based on pre-trained Vision Transformer~\cite{dosovitskiy2021vit}.
Regrettably, methods based on pre-trained models have not sufficiently addressed
the critical challenge of non-IID issue in federated learning~\cite{zhao2018federated,li2022federated},
lacking developed methods to mitigate resulting performance degradation effectively.
In federated learning, the non-IID nature introduces data heterogeneity, leading to significant bias in models trained on different clients,
and aggregating these biased models directly may result in a suboptimal global model~\cite{li2022federated,zhang2022federated}.
Similar to conventional federated learning, FCL encounters performance degradation with increasing non-IID degrees, as outlined in~\cite{zhang2023target}.
Additional knowledge sharing among clients is necessary to mitigate the impact of non-IID.

To tackle the issue of non-IID, lightweight prototypes~\cite{snell2017proto},
defined as the mean of extracted features from data,
have been employed as efficient and privacy-preserving information carriers in federated learning~\cite{tan2022fedpcl,huang2023rethinking}.
Building on the insights from~\cite{tan2022fedpcl,huang2023rethinking},
we propose using prototypes as a superior alternative to rehearsal for addressing performance degradation caused by both catastrophic forgetting and non-IID issues.
To reduce the communication cost associated with transmitting pre-trained transformers,
we draw inspiration from prompt-based continual learning methods~\cite{wang2022l2p, wang2022dualprompt, smith2023coda-prompt},
which utilize lightweight prompts~\cite{lester2021power,li2021prefix,jia2022vpt} to instruct models.

In this work, an efficient and non-IID robust federated continual learning framework,
called {\em{\textbf{F}ederated \textbf{P}rototype-Augmented \textbf{P}rompt \textbf{L}earning (FPPL)}}, is proposed.
The proposed FPPL leverages both prompt tuning and prototypes to tackle the challenges of non-IID and catastrophic forgetting.
Specifically, task-specific prompts are integrated with a fusion function to fully leverage acquired knowledge,
thereby mitigating catastrophic forgetting.
To address performance degradation caused by both non-IID data and catastrophic forgetting,
a classifier debiasing mechanism based on local prototypes is designed.
These local prototypes are aggregated into global prototypes on the server and returned to clients,
achieving unified representations across clients through contrastive learning and further addressing the non-IID issue.
The main contributions can be summarized as:

\noindent - The prompt tuning based on pre-trained transformers is introduced into FCL, eliminating the need for rehearsal and reducing communication cost;

\noindent - The prototypes are employed to address the non-IID issue in FL through lightweight transmission while stabilizing local training and mitigating catastrophic forgetting globally;

\noindent - Extensive experiments on ImageNet-R, CUB-200 and CIFAR-100 can validate the effectiveness of FPPL, achieving notable performance with strong non-IID robustness.

\section{Related Work}
\label{sec:related_work}

\subsection{Federated Continual Learning}

Federated Learning (FL) is a distributed learning paradigm that enables clients to collaboratively train a global model without sharing their local private data~\cite{mcmahan2017communication,yang2019federated,kairouz2021advances,huang2024federated}.
However, this paradigm often assumes that clients' private data remains static, which is inconsistent with the dynamic nature of data accumulation over time.

Federated continual learning (FCL)~\cite{yoon2021federated,dong2022fcil} aims to address the challenge
of clients acquiring knowledge across a sequence of tasks by incorporating continual learning techniques into the federated learning setting.
FCL confronts challenges including catastrophic forgetting~\cite{kirkpatrick2017overcoming,rebuffi2017icarl} in continual learning
and non-IID issue~\cite{zhao2018federated,li2022federated} in federated learning.

The study of FCL has gained substantial traction in recent years.
FedWeIT~\cite{yoon2021federated} utilizes decomposed global federated parameters and sparse task-specific parameters
to minimize interference between incompatible tasks and allow positive knowledge transfer across clients.
GLFC~\cite{dong2022fcil} designs class-aware and class-semantic loss for FCL, and utilizes a proxy server to reconstruct local data.
Approaches~\cite{qi2023better,zhang2023target} leverage generative models to generate synthetic data for knowledge distillation~\cite{hinton2015distilling} to alleviate catastrophic forgetting.

With the development of pre-trained models~\cite{han2021pre},
FCL methods based on pre-trained transformers~\cite{vaswani2017attention,devlin2019bert,dosovitskiy2021vit} have achieve notable performance.
FedET~\cite{liu2023FedET} incorporates enhancer modules to address catastrophic forgetting challenge while reducing communication overhead.
Fed-CPrompt~\cite{bagwe2023fedcprompt}, based on prompts, facilitates communication-efficient FCL without the need for rehearsal.

While FCL methods using pre-trained transformers have demonstrated notable performance, their effectiveness across varying non-IID degrees remains underexplored.

\subsection{Continual Learning}

Continual learning, alternatively known as incremental learning or lifelong learning~\cite{de2022continual,masana2023class,wang2024comprehensive,zhou2024continual},
is a paradigm involving models acquiring knowledge from sequential tasks.
Throughout the learning process of new tasks, models are either restricted from accessing or can only partially access data pertaining to prior tasks.
The primary goal of continual learning is to mitigate the phenomenon of forgetting previous tasks while concurrently acquiring knowledge on new ones.
Continual learning can be broadly categorized into three classes: regularization-based methods, rehearsal-based methods, and model expansion-based methods.

Regularization based methods aim to constrain the adjustment of model parameters during the acquisition of new tasks~\cite{kirkpatrick2017overcoming,li2018lwf,smith2023closer}.
Rehearsal based methods such as iCaRL~\cite{rebuffi2017icarl} and BiC~\cite{wu2019bic},
involve the rehearsal of historical data from previous tasks during the learning of new tasks.
Model expansion-based methods, like DER~\cite{yan2021der} and MEMO~\cite{zhou2023memo},
introduce additional network structures during the acquisition of new tasks.

Recent advancements in continual learning include methods that leverage knowledge from pre-trained transformers.
L2P~\cite{wang2022l2p} introduced a key-query matching mechanism for selecting prompts from a prompt pool to instruct the knowledge from the pre-trained model for learning new tasks.
Building upon L2P, DualPrompt\cite{wang2022dualprompt} enhanced performance by incorporating both general prompts and expert prompts.
CODA-Prompt~\cite{smith2023coda-prompt} utilizing an attention mechanism to construct input-conditioned prompts.

\section{Preliminaries}
\label{sec:pre}

In FCL~\cite{yoon2021federated,dong2022fcil},
a server-coordinated set of $K$ clients sequentially learns over $T$ tasks.
Training data are randomly distributed among clients and cannot be uploaded.
To emphasis, the challenge of class-incremental learning~\cite{rebuffi2017icarl} in continual learning is mainly focused in this work.
Class-incremental learning involves $T$ sets of non-overlapping class-labeled training datasets, denoted as
$\{ \mathcal{D}^1, \cdots, \mathcal{D}^T\ :\ \mathcal{D}^t = {(x_i^t, y_i^t)}_{i=1}^{n^t} \}$,
where $\mathcal{D}^t$ represents the $t$-th training dataset, and $y_i^t \in \mathcal{Y}^t$ denotes the corresponding label of $x_i^t$.
For any two distinct datasets, the intersection of their class categories is empty,
which can be expressed as $\mathcal{Y}^{t} \cap \mathcal{Y}^{t'} = \emptyset$, for any ${t} \neq {t'}$.
During the $t$-th task phase, updates are only permitted by employing $\mathcal{D}^t$
or, additionally, by incorporating a small amount of rehearsal~\cite{rebuffi2017icarl}.
The new obtained model needs to classify all known classes in the first $t$ stages $\cup_{i=1}^{t} \mathcal{Y}^i$.

Furthermore, by expanding the setting of class-incremental learning to federated learning scenario,
the $t$-th task's training data $\mathcal{D}^t$ is distributed randomly among $K$ clients as
$\mathcal{D}^t = \{{\mathcal{D}_k^t} \}_{k=1}^{K}$.
It is assumed that the data heterogeneity across clients are distribution-based label skew~\cite{li2022federated,zhang2022federated},
and for any clients $k \neq k'$, it holds that $\mathcal{D}_{k}^t \cap \mathcal{D}_{k'}^t = \emptyset$.

\section{Methodology of the FPPL}
\label{sec:method}

With the advancement of pre-trained models~\cite{han2021pre},
prompt-based continual learning methods~\cite{wang2022l2p,wang2022dualprompt,smith2023coda-prompt} have achieved notable performance without rehearsal.
However, these methods are tailored for scenarios within a centralized dataset and it is not obvious to be directly applied to FL.
In addition, due to the non-IID (non-independent and identically distributed) issue on client datasets~\cite{zhao2018federated,li2022federated},
models trained locally may exhibit strong biases,
and directly aggregating these models might not satisfy the desired global requirements~\cite{li2022federated,zhang2022federated}.
Furthermore, the global aggregated model returned to local clients might overfit to local biases, and potentially lead to knowledge forgetting of aggregated model and slow convergence speed~\cite{huang2022learn}.

To address this challenge, the training and implementing procedures of both the clients and the server need to be properly established:
1) For each client, local prototypes are extracted and simultaneously transferred to the server.
The client will also utilize the global prototypes from the server to update local model
through contrastive learning~\cite{chen2020simclr,he2020moco}, achieving a global knowledge-smoothed convergence;
2) For the server, it first aims to aggregate collected prototypes from clients, which further are send back to all clients.
The locally uploaded prototypes from each client will also be leveraged for lightweight classifier debiasing,
with the purpose to address biases arising from missing historical tasks and non-IID-induced bias in local models.

\subsection{Local Prompt Tuning}
Motivated by some recent works~\cite{wang2022l2p,wang2022dualprompt,smith2023coda-prompt},
the pre-trained Vision Transformer (ViT)~\cite{dosovitskiy2021vit,vaswani2017attention}
is introduced with efficient prompt tuning~\cite{lester2021power,jia2022vpt} based on clients' local data.
Task-specific prompts are employed for each learning task inspired by previous work~\cite{wang2022dualprompt},
while freezing the learned prompts at the end of each task to prevent catastrophic forgetting.
A fusion function is utilized to combine all task-specific prompts and maximize the utilization of accumulated knowledge.

For prompt tuning, a set of prompts $ p = \{ p^{(l)} \}_{l=start}^{end}$,
where $ p^{(l)} \in \mathbb{R}^{L_p \times D}$, will be utilized to instruct pre-trained ViT.
The terms $start$ and $end$ denote the corresponding index of the layers in the ViT Transformer block where prompts are inserted,
$ L_p $ represents the length of prompt parameters, and $D$ indicates the embedding dimension of ViT.
Following the insertion layers in~\cite{wang2022dualprompt}, where $start = 1$ and $end = 5$,
prompts are appended to encoded features at specific layers of the ViT Transformer block.
For convenience, we use $\left[ \mathbf{x} , p \right]$ to denote the combination of the input $\mathbf{x}$ with the prompt $p$.

\begin{figure}[htp!]
    \centering
    \includegraphics[width=0.99\linewidth]{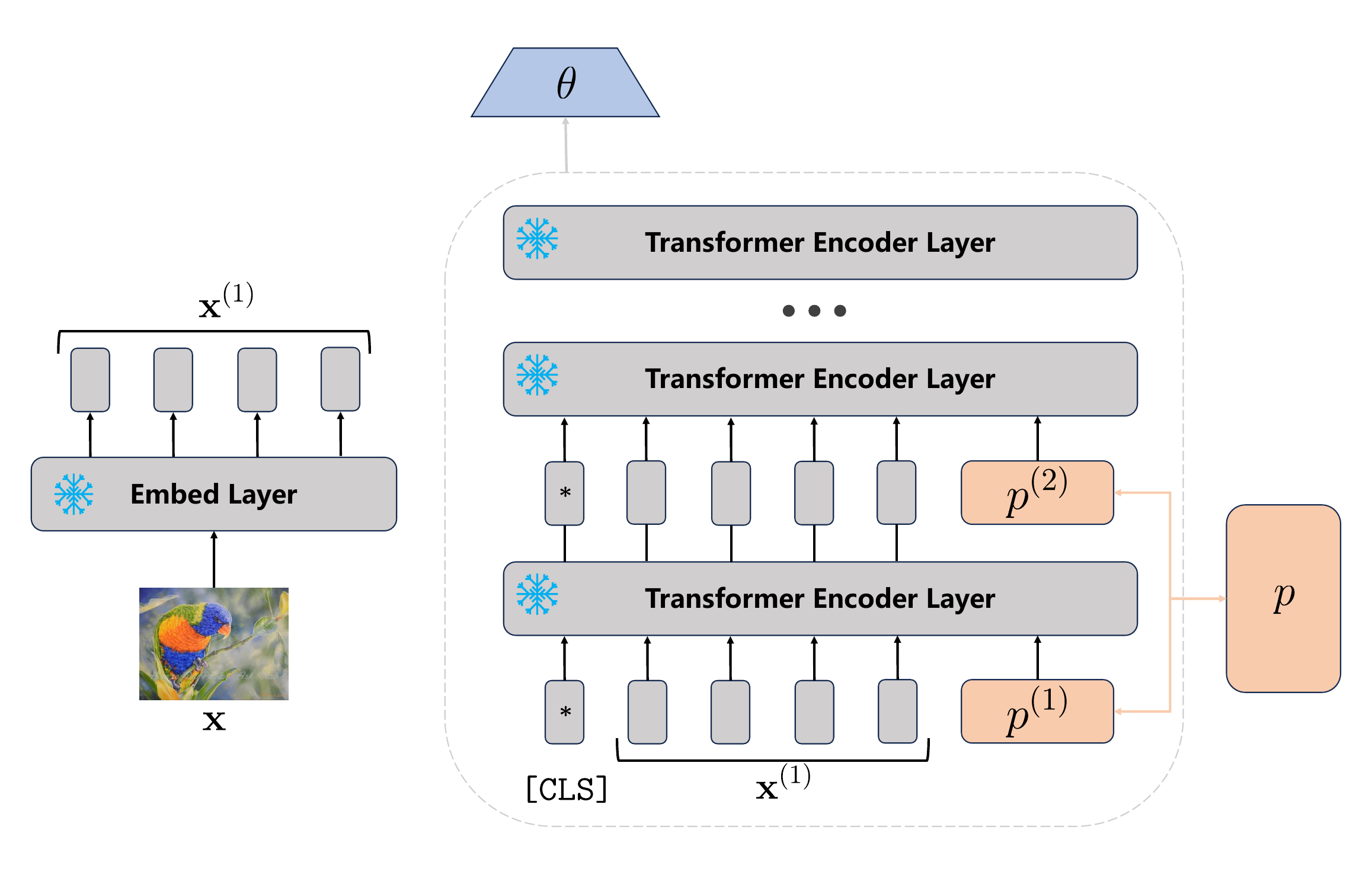}
    \caption{
        Illustration of the prompt tuning utilized in our FPPL.
        The lightweight prompt and classifier parameters are tuned during learning,
        while achieving parameter-efficient tuning for pre-trained Vision Transformer.
    }
    \label{fig:prompt_tuning}
\end{figure}

During the learning process, the backbone parameters $\Phi$ of ViT remain frozen,
while only the task-specific prompt $p_t$, coslinear parameters $\psi$, and classifier parameters $\theta$ are properly tuned.
These parameters are collectively expressed as $\bm{w} = \left\{ \psi, p_t, \theta \right\}$.
At the first task ($t=1$), the task-specific prompt $p_1$ is uniformly initialized and then optimized.
After completing a task ($t \leftarrow t+1$), the newly obtained task-specific prompt is initialized as the previous prompt ($p_{t+1} \leftarrow p_t$) to obtain acquired knowledge.

To mitigate catastrophic forgetting, the prompts corresponding to previously completed tasks $\{ p_i \}_{i=1}^{t-1}$ will be frozen following~\cite{wang2022dualprompt}.
At task $t$, we will gather $P = \{ p_i \}_{i=1}^t$, where $p_t$ denoting the $t$-th task-specific prompt.
Since the newly obtained prompt is based on previous prompts,
we utilize a tunable coslinear to derive weights for task-specific prompts.
These weights are further integrated through a fusion function to fully exploit accumulated knowledge.

\begin{figure*}[htp!]
    \centering
    \includegraphics[width=0.95\textwidth]{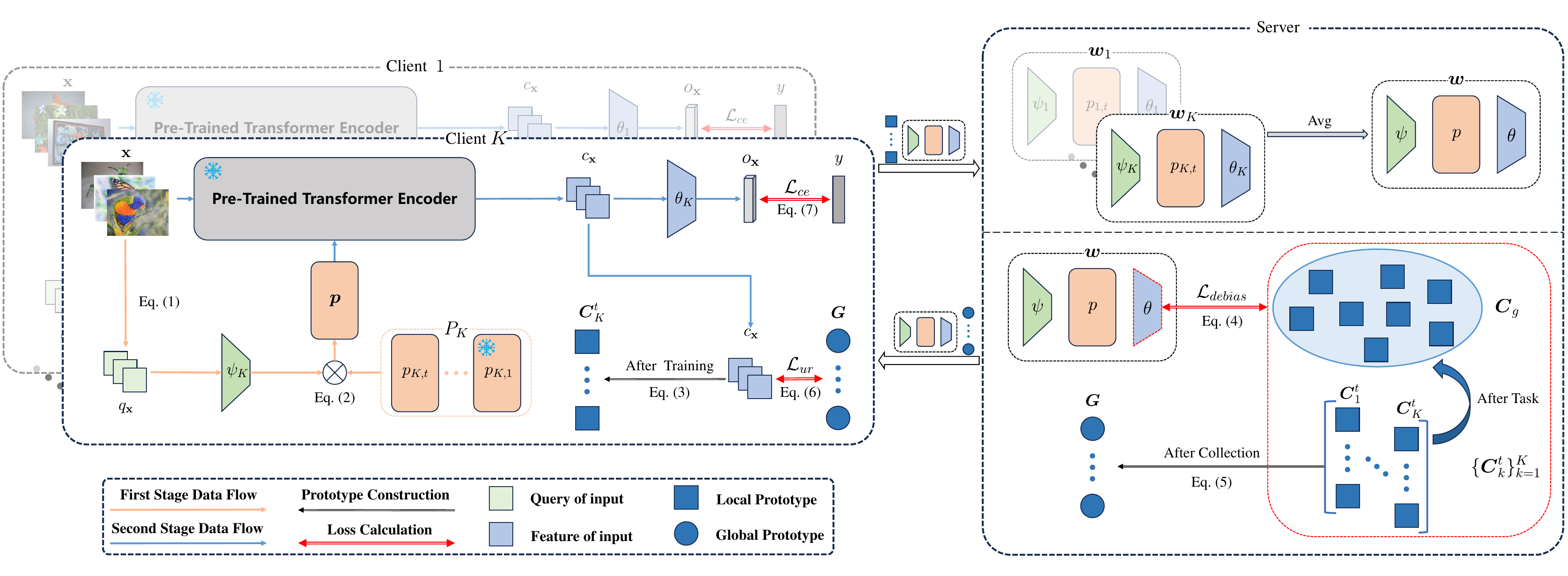}
    \caption{
        The overview of proposed FPPL framework.
        1) Client: Extract unified representations across different clients by $\mathcal{L}_{ur}$
        and construct local prototype set using Eq. \eqref{eq:prototype_local} after training local model.
        2) Server: Debias the classifier of the aggregated parameters by $\mathcal{L}_{debias}$
        and construct global prototypes returned to all clients using Eq. \eqref{eq:prototype_global}.
        3) Lightweight prompts, linear parameters and prototypes are transmitted between clients and the server,
        while no additional data is stored on each client.
    }
    \label{fig:overview}
\end{figure*}

Utilizing the whole pre-trained model as a frozen feature extractor following~\cite{wang2022l2p}, query is denoted as:
\begin{equation}\label{eq:function_query}
    q_{\mathbf{x}} = f_{\Phi} \left( \mathbf{x} \right).
\end{equation}

The query is further processed by the coslinear layer $g_{\psi}$,
mapping the query to a $t$-dimensional output based on cosine similarity,
i.e., $g_{\psi}(q_{\mathbf{x}}) = \frac{q_{\mathbf{x}} \psi} {\|q_{\mathbf{x}}\|_{2} \|\psi\|_{2}} \in \mathbb{R}^t$.
The weights for each task-specific prompt are computed using the softmax function,
and an element-wise multiplication operation is applied to derive the following fused prompt:
\begin{equation}\label{eq:function_fusion}
    {\textstyle{\boldsymbol{p} = \sum_{i=1}^{t} \left( \sigma \left( g_{\psi}(q_{\mathbf{x}}) \right) \odot P \right)_i,}}
\end{equation}
where $\sigma (\cdot)$ denotes the softmax function and $\odot$ denotes the element-wise multiplication operation.

The complete feature extraction function is then denoted as
$c_{\mathbf{x}} = f_{\Phi} \left( \left[ \mathbf{x}, \boldsymbol{p} \right] \right)$,
in which $\left[ \mathbf{x}, \boldsymbol{p} \right]$ is the combination of input and fused prompt.
Thus the final output logits is denoted as $o_{\mathbf{x}} = h_{\theta}(c_{\mathbf{x}})$ for classification.

For brevity, the query, extracted feature, and final output logits of input $\mathbf{x}$ are denoted as $q_{\mathbf{x}}$, $c_{\mathbf{x}}$, and $o_{\mathbf{x}}$, respectively.

Thanks to the lightweight nature of prompt parameters, it becomes feasible to train and store a set of prompt parameters for each task.
During inference, the weights are assigned to task-specific prompts based on the similarity with query, and all prompt parameters are fused to fully utilize the accumulated knowledge.
This combination of advantages makes the local prompt tuning approach efficient and scalable.

\subsection{Global Prototype Aggregation}
The local prompt tuning can alleviate catastrophic forgetting locally,
however the heterogeneity of data across clients can lead to biased models,
and direct aggregation may potentially yield suboptimal performance~\cite{li2022federated,zhang2022federated}.
Motivated by~\cite{tan2022fedpcl,huang2023rethinking},
FPPL employs prototypes ~\cite{snell2017proto} to fully utilize the information from each client
to address this issue and further mitigate catastrophic forgetting globally.
The local prototype is obtained by averaging features belonging to the same class:
\begin{equation}\label{eq:prototype_local}
    {\textstyle{\mathcal{C}_k^y = \frac{1}{| \mathcal{D}_k^y |} \sum_{(\mathbf{x}, y) \in \mathcal{D}_k^y} c_{\mathbf{x}},}}
\end{equation}
where $c_{\mathbf{x}}$ is the extracted feature of input $\mathbf{x}$
and $\mathcal{D}_k^y$ denotes the subset of private dataset of client $k$ with label $y$.

After local training procedure in current round,
the obtained model will extract local prototypes to form local prototype set
$\boldsymbol{C}_k^t = \{\mathcal{C}_k^y\}_{y \in \mathcal{Y}_k^t}$
and $\mathcal{Y}_k^t$ represents the available classes of client $k$ at task $t$.
These prototype sets are synchronously transferred to the server.
Further, these locally uploaded prototype sets can be merged on the server with the prototype pool as
$\boldsymbol{C}_{g} = \boldsymbol{C}_{g} \cup \{ \boldsymbol{C}_k^t \}_{k=1}^{K}$,
after the completion of current task $t$, to retain knowledge from all tasks.

The classifier often exhibits strong bias in continual learning~\cite{wu2019bic,zhang2023slca}
which could be exacerbated by the non-IID issue of decentralized local datasets~\cite{li2022federated,zhang2022federated}.
This leads us to employ the uploaded local prototypes for classifier debiasing on the server,
the loss of the server is then established as:
\begin{equation}\label{eq:l_debias}
    {\textstyle{\mathcal{L}_{debias} \left( \theta \right) = \sum_{\mathcal{C}_k^y \in \boldsymbol{C}_{g} \cup \{ \boldsymbol{C}_k^t \}_{k=1}^{K}} \ell_{ce} \left( y, h_{\theta}(\mathcal{C}_k^y) \right),}}
\end{equation}
where $\ell_{ce}(\cdot)$ denotes the cross entropy loss function.

It is worth emphasizing that the debias operation on the server side does not cause excessive computational and storage pressure,
because of the lightweight of the prototype collection and parameters.
The effectiveness of this debias operation for performance degradation alleviation can be demonstrated in the following experiment section.

To consolidate these local information carriers into global information
and further mitigate the non-IID issue in FCL, the global prototype is established as:
\begin{equation}\label{eq:prototype_global}
    {\textstyle{\mathcal{G}^y = \frac{1}{|\mathcal{K}^y|} \sum_{k \in \mathcal{K}^y} \mathcal{C}_k^y,}}
\end{equation}
where $ \mathcal{K}^y $ denotes the set of clients containing class $ y $ data.
The global prototype set $\boldsymbol{G} = \{ \mathcal{G}^y \}_{y \in \mathcal{Y}^t}$ is feedback to all clients
with the purpose to obtain unified representations across clients.
Contrastive learning~\cite{chen2020simclr,he2020moco,li2021moon} is utilized to unify the representations across clients through the following unified representation loss:
\begin{equation}\label{eq:l_ur}
    {\textstyle{\mathcal{L}_{ur}(p_t, \psi) = \sum_{(\mathbf{x},y) \in \mathcal{D}_k^t} - \log \frac{\exp \left(\textrm{sim}(c_{\mathbf{x}}, \mathcal{G}^{y}) / \tau\right)}{\sum_{y_a \in \mathcal{Y}^t} \exp \left(\textrm{sim}(c_{\mathbf{x}}, \mathcal{G}^{y_a}) / \tau\right)},}}
\end{equation}
where $\mathcal{Y}^t$ represents the global available classes at task $t$ and $c_{\mathbf{x}}$ is the extracted feature of input $\mathbf{x}$.
The function $\textrm{sim}(\cdot, \cdot)$ denotes the cosine similarity function following~\cite{chen2020simclr,li2021moon}.
$\tau$ denotes the temperature that controls the tolerance of difference between extracted feature and corresponding global prototype.

It is worth emphasizing that the size of the returned global prototype aligns with the local prototype generated by each client, ensuring that no extra transmission burden is introduced during the process.

To summarize the training process on each client,
the cross-entropy loss for classification is expressed as:
\begin{equation}\label{eq:l_ce}
    {\textstyle{\mathcal{L}_{ce}(p_t, \psi, \theta) = \sum_{(\mathbf{x},y) \in \mathcal{D}_k^t} \ell_{ce}(y, o_{\mathbf{x}}),}}
\end{equation}
where $o_{\mathbf{x}}$ is the final output logits of input $\mathbf{x}$ and $y$ denotes the label of input $\mathbf{x}$.

The overall loss function combines the unified representation loss with the classification loss, yielding:
\begin{equation}\label{eq:l_client}
    \mathcal{L}_{client}(p_t, \psi, \theta) = \mathcal{L}_{ur}(p_t, \psi) + \mathcal{L}_{ce}(p_t, \psi, \theta).
\end{equation}

\section{Experiments}
\label{sec:exp}

\subsection{Experimental Setup}

\noindent\textbf{Datasets and Non-IID Setting.}
We utilize three datasets, ImageNet-R~\cite{hendrycks2021many}, CUB-200~\cite{wah2011caltech}, and CIFAR-100~\cite{krizhevsky2009learning}.
The CIFAR-100 dataset contains 100 different classes of natural images, each with 500 training samples and 100 testing samples.
The CUB-200 dataset includes 200 subcategories of fine-grained bird images, totaling about 12,000 samples.
The ImageNet-R dataset consists of 200 classes, containing 24,000 training samples and 6,000 testing samples.
The ImageNet-R and CUB-200 datasets are randomly split into 20 tasks $T=20$, each containing an equal number of disjoint classes,
while the CIFAR-100 dataset is divided into 10 tasks $T=10$.

Notably, all images within ImageNet-R are considered out-of-distribution samples,
despite having overlapping classes with ImageNet~\cite{deng2009imagenet}, the dataset used for pre-training.
These images represent challenging instances or newly acquired data characterized by diverse stylistic attributes.
Consequently, ImageNet-R is a robust metric for evaluating the generalization capability of pre-trained models,
which is utilized in assessing performance in continual learning methods~\cite{wang2022dualprompt,smith2023coda-prompt,zhang2023slca}.

To simulate non-IID setting, we employ the Dirichlet distribution, which is widely used in FL~\cite{li2022federated,zhang2022federated}.
The parameter $\beta$ regulates the extent of non-IID, with a lower value of $\beta$ indicating a higher degree of non-IID.

\noindent\textbf{Evaluation Metric.}
In the evaluation of method performance, we employ the metrics of accuracy $A_t$,
average accuracy $\bar{A}$ and average forgetting $\bar{F}$, as established in~\cite{chaudhry2018riemannian}
and widely adopted in continual learning ~\cite{wang2022l2p,wang2022dualprompt,smith2023coda-prompt}.

The accuracy on task $i$ after learning task $t$ is denoted as $a_{i, t}$,
and $A_t = \frac{1}{t} \sum_{i=1}^{t} a_{i, t}$ is defined as the accuracy of all tasks after learning task $t$.
The overall performance after task $T$ is quantified through the calculation of average accuracy $\bar{A} = \frac{1}{T} \sum_{t=1}^{T} A_{t}$,
where $T$ signifies the total number of tasks, thus providing a comprehensive measure across the entire learning progression.

To measure the degree of forgetting, the average forgetting is calculated as,
$\bar{F} = \frac{1}{T-1} \sum_{i=1}^{T-1} \max_{t \in \{ 1, \cdots, T-1 \}} (a_{i, t} - a_{i, T})$.
Here, $\max_{t \in \{ 1, \cdots, T-1 \}} (a_{i, t} - a_{i, T})$
mesures the gap between the accuracy of task $i$ at task $T$ and the optimal accuracy achievable for task $i$,
capturing the forgetting of previous tasks after the whole learning process.

\begin{algorithm}[htbp] 
	\renewcommand{\algorithmicrequire}{\textbf{Input:}}
	\renewcommand{\algorithmicensure}{\textbf{Output:}}
	\caption{The FPPL Framework} 
	\label{alg1} 
    {\bf Input:}
    Task sequence length $T$, communication rounds per task $R$, number of clients $K$, 
    local epochs $E$, server epochs $E_{server}$, private datasets $\{ \mathcal{D}_k^t \}_{k=1}^{K}$.
	\begin{algorithmic}[1]
    \Procedure{FPPL}{}
        \State 
        Initialize $\bm{w} = \{ \psi, p, \theta \}$, 
        global prototype set $\boldsymbol{G} = \{\}$, and prototype pool $\boldsymbol{C}_g = \{\}$;
        \For{$t = 1, 2, \cdots, T$}
            \For{$r = 1, 2, \cdots, R$}
                \For{$k = 1, 2, \cdots, K$ {\bf in parallel}}
                    \State $\bm{w}_{k}^{r+1}, \boldsymbol{C}_{k}^t \leftarrow $ \textproc{Client} $(t, r, k, {\bm{w}}^{r}, {\boldsymbol{G}})$;
                \EndFor
                \State $\bm{w}^{r+1} \leftarrow \sum_{k=1}^K \frac{| \mathcal{D}_k^t |}{| \mathcal{D}^t |} \bm{w}_{k}^{r+1}$;
                \For{$e = 1, 2, \cdots, E_{server}$}
                    \State Optimize $\theta^{r+1}$ by Eq. \eqref{eq:l_debias};
                \EndFor
                \State $\boldsymbol{G} = \{ \mathcal{G}^y \}_{y \in \mathcal{Y}^t}$ by Eq. \eqref{eq:prototype_global};
            \EndFor
            \State $\boldsymbol{C}_g \leftarrow \boldsymbol{C}_g \cup \{{\boldsymbol{C}}_k^t\}_{k=1}^{K}$;
        \EndFor
    \EndProcedure
    \Statex
    \Function{Client}{$t, r, k, {\bm{w}}, {\boldsymbol{G}}$}
        \State $\bm{w}_k = \{ \psi_k, p_{k, t}, \theta_k \} \leftarrow \bm{w}$;
        \If{$r = 1$ and $t > 1$}
            \State $p_{k, t-1} \leftarrow p_{k, t}$;
            \State Freeze $p_{k, t-1}$;
        \EndIf
        \For{$e = 1, 2, \cdots, E$}
            \For{$(\mathbf{x}, y)$ in $\mathcal{D}_k^t$}
                \State Compute $\mathcal{L}_{ur}$ by Eq. \eqref{eq:l_ur} with $(c_{\mathbf{x}}, \boldsymbol{G})$;
                \State Compute $\mathcal{L}_{ce}$ by Eq. \eqref{eq:l_ce} with $(o_{\mathbf{x}}, y)$;
                \State Optimize $\bm{w}_k$ by Eq. \eqref{eq:l_client} with $\mathcal{L}_{ur}, \mathcal{L}_{ce}$;
            \EndFor
        \EndFor
        \State Local prototype set $\boldsymbol{C}_k^t = \{\mathcal{C}_k^y\}_{y \in \mathcal{Y}_k^t}$ by Eq. \eqref{eq:prototype_local};
        \State \textbf{return} $\bm{w}_k, \boldsymbol{C}_k^t$.
    \EndFunction
\end{algorithmic} 
\end{algorithm}

Additionally, we calculate the parameter transmission cost per communication round as an additional metric,
measuring the communication efficiency of different methods, denoted as $COMM$.

\noindent\textbf{Baselines.}
We compare our approach with both rehearsal-free continual learning methods and rehearsal-based federated continual learning method:

\noindent - Regularization-based methods EWC~\cite{kirkpatrick2017overcoming} and LwF~\cite{li2018lwf};

\noindent - SLCA~\cite{zhang2023slca}, an advanced continual learning method using slow learner and classifier alignment;

\noindent - Prompt-based continual learning methods using pre-trained model: L2P~\cite{wang2022l2p}, DualPrompt~\cite{wang2022dualprompt}, and CODA-Prompt~\cite{smith2023coda-prompt};

\noindent - GLFC~\cite{dong2022fcil}, a rehearsal-based federated continual learning method, stores selected samples on the client side and reconstructs local data on the server side.

\begin{table*}[htp!]
    \caption{
        \textbf{Results on 20-task ImageNet-R and 20-task CUB-200}.
        $\bar{A}$ gives the accuracy averaged over tasks,
        $A_{T}$ gives the accuracy of all tasks after the last incremental stage,
        $\bar{F}$ gives the average forgetting,
        and $COMM$ gives the communication cost per round.\\
        \centering We report results over 3 trials.
    }
    \label{tab:overallperformance}
    \centering

    \scalebox{0.99}{
        \begin{tabular}{c|c c c||c c c|c}
            \hline
            \multirow{2}{*}{Method} & \multicolumn{3}{c||}{20-task ImageNet-R} & \multicolumn{3}{c|}{20-task CUB-200} & \multirow{2}{*}{$COMM$ ($\downarrow$)} \\
            \cline{2-7}
            \rule{0pt}{10pt}        & $\bar{A}$ ($\uparrow$)                   & $A_{T}$ ($\uparrow)$                 & $\bar{F}$ ($\downarrow$)
                                    & $\bar{A}$ ($\uparrow$)                   & $A_{T}$ ($\uparrow)$                 & $\bar{F}$ ($\downarrow$)               \\
            \hline
            FedEWC
                                    & $69.20 \pm 0.90$                         & $55.60 \pm 0.58$                     & $24.19 \pm 1.49$
                                    & $63.93 \pm 0.68$                         & $46.92 \pm 1.84$                     & $26.70 \pm 3.61$
                                    & $85.95\times 10^6$                                                                                                       \\
            FedLwF
                                    & $41.88 \pm 1.31$                         & $19.03 \pm 2.21$                     & $49.75 \pm 1.71$
                                    & $30.17 \pm 7.59$                         & $12.16 \pm 3.81$                     & $68.06 \pm 9.44$
                                    & $85.95\times 10^6$                                                                                                       \\
            FedL2P
                                    & $48.83 \pm 4.77$                         & $41.37 \pm 2.62$                     & $\bm{2.40 \pm 1.38}$
                                    & $58.14 \pm 5.19$                         & $41.55 \pm 1.07$                     & $\bm{6.88 \pm 1.88}$
                                    & $\bm{00.20\times 10^6}$                                                                                                  \\
            FedDualP
                                    & $56.17 \pm 1.84$                         & $48.58 \pm 1.01$                     & $9.36 \pm 0.82$
                                    & $66.16 \pm 2.22$                         & $49.63 \pm 0.65$                     & $14.18 \pm 1.44$
                                    & $00.26\times 10^6$                                                                                                       \\
            FedCODA-P
                                    & $59.56 \pm 2.76$                         & $52.40 \pm 1.61$                     & $7.29 \pm 0.83$
                                    & $65.58 \pm 1.93$                         & $47.86 \pm 1.26$                     & $13.17 \pm 2.16$
                                    & $00.35\times 10^6$                                                                                                       \\
            FedSLCA
                                    & $71.42 \pm 0.77$                         & $56.26 \pm 1.12$                     & $34.60 \pm 1.41$
                                    & $76.62 \pm 1.93$                         & $60.47 \pm 2.08$                     & $35.23 \pm 1.71$
                                    & $91.86\times 10^6$                                                                                                       \\
            GLFC
                                    & $73.64 \pm 0.46$                         & $56.42 \pm 0.06$                     & $33.95 \pm 0.62$
                                    & $84.81 \pm 1.22$                         & $71.52 \pm 1.06$                     & $19.14 \pm 0.67$
                                    & $87.57\times 10^6$                                                                                                       \\
            \textbf{FPPL}
                                    & $\bm{74.33 \pm 0.37}$                    & $\bm{70.84 \pm 0.15}$                & $5.07 \pm 0.27$
                                    & $\bm{88.80 \pm 0.50}$                    & $\bm{80.93 \pm 1.02}$                & $13.29 \pm 1.22$
                                    & $00.25\times 10^6$                                                                                                       \\
            \hline
        \end{tabular}
    }
\end{table*}

Recent findings by SLCA~\cite{zhang2023slca} suggest that the performance of full fine-tuning methods can be enhanced
by utilizing a lower learning rate for the representation layer of the pre-trained model and a slightly higher rate for the classifier, referred to as the slow learner approach.
Consequently, we enhance EWC, LwF and GLFC~\footnote{
    The original paper~\cite{dong2022fcil} utilizes ResNet18~\cite{he2016resnet} as the backbone.
    For fairness, we replace it with the ViT-B/16 model pre-trained on ImageNet, aligning with other methods~\cite{wang2022l2p,wang2022dualprompt,smith2023coda-prompt}.
} with the slow learner.

To emphasis, we will implement these continual learning methods in the FCL scenario,
while naming them as FedEWC, FedLwF, FedL2P, FedDualP, FedCODA-P, and FedSLCA, respectively.

\noindent\textbf{Implementation Details.}\phantomsection\label{impdetails}
Consistent with prior works~\cite{wang2022l2p,wang2022dualprompt,smith2023coda-prompt},
we employ a ViT-B/16 model pre-trained on ImageNet as the common backbone for all baseline methods.
Adam~\cite{kingma2014adam} optimizer is used for all methods with the same batch size of 128.
Prompt-based methods uniformly use a constant learning rate of $lr = 1e^{-3}$.
Conversely, full fine-tuning methods use a learning rate of $lr_{rps} = 1e^{-4}$ for the representation layer
and $lr_{cls} = 1e^{-2}$ for the classifier, in accordance with the slow learner proposed by SLCA~\cite{zhang2023slca}.
We set clients $K=10$, overall communication rounds $R_{all}=200$ and local update epochs $E = 5$,
the communication rounds of each task is calculated by $R = {\textstyle{\frac{R_{all}}{T}}}$ following ~\cite{dong2022fcil,zhang2023target}.
For the FPPL, server update is performed over $E_{server}=5$ and temperature is set to $\tau = 0.2$ empirically.

GLFC maintains an exemplary memory of $\mathcal{M}_l = 2000$ on each client for rehearsal.
In the original paper~\cite{dong2022fcil}, the private datasets of each client overlap,
making their volume approach the centralized dataset ($60 \%$).
In contrast, our setting has non-overlapping datasets, as described in Section~\ref{sec:pre},
so each client's dataset is only $10 \%$ of the centralized dataset when $K = 10$.
With $\mathcal{M}_l = 2000$, the overall exemplary memory is $\mathcal{M} = \mathcal{M}_l \times K = 20,000$,
which exceeds the volume of the CUB-200 dataset and approaches the size of the ImageNet-R dataset,
making it nearly consistent with training in static scenarios.
Therefore, we set exemplary memory $\mathcal{M}_l = \frac{2000}{K} = 200$ on each client.

All experiments are conducted across three random seeds 2023, 2024, and 2025,
with the final results derived from the mean across these runs.

\subsection{Performance Comparison}

\begin{figure*}[ht]
    \subfigure[Accuracy curves on 20-task ImageNet-R (left) and 20-task CUB-200 (right).]{
        \begin{minipage}[t]{0.58\linewidth}
            \centering
            \includegraphics[width=0.99\linewidth]{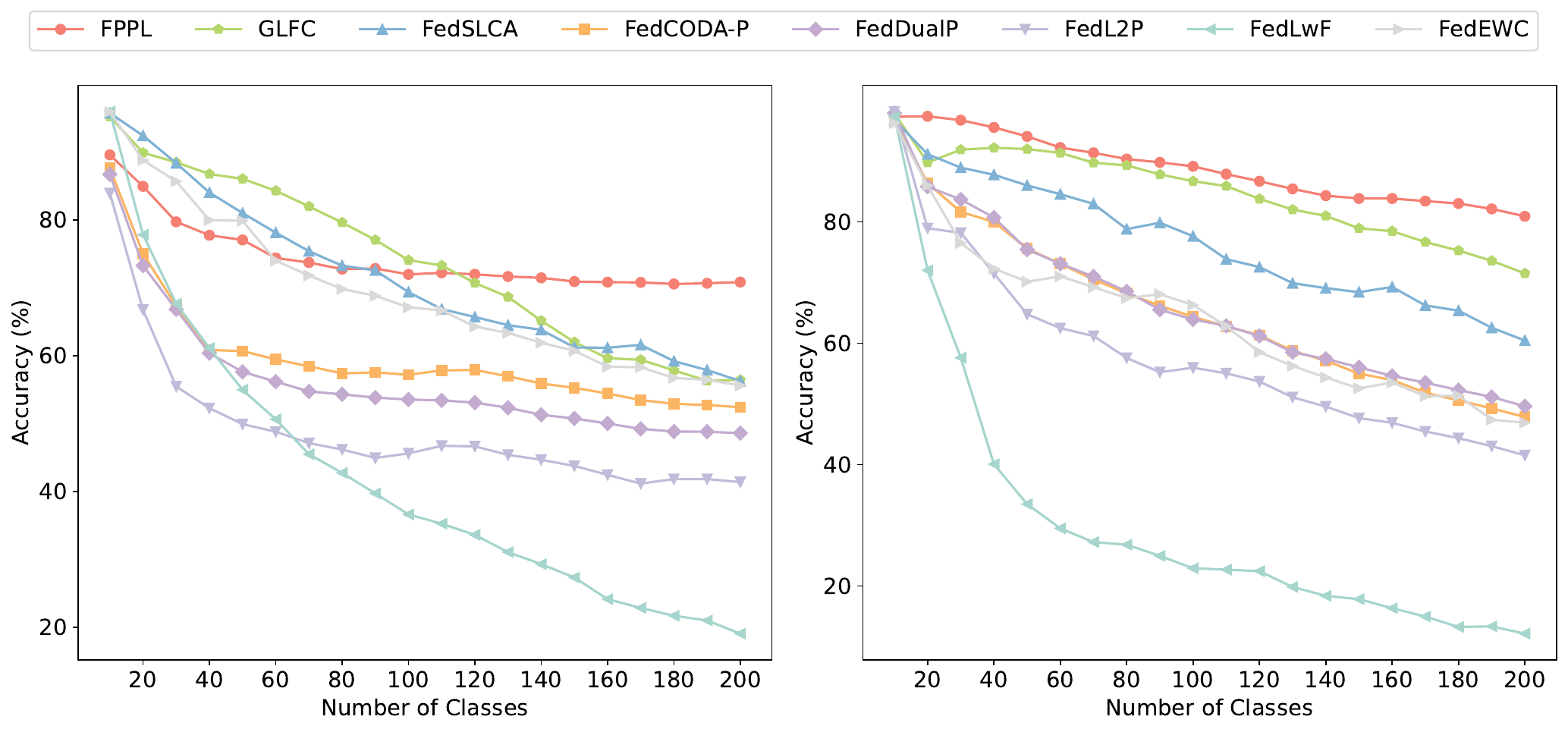}
            \label{fig:incacc}
        \end{minipage}
    }
    \subfigure[Average accuracy on 10-task CIFAR-100.]{
        \begin{minipage}[t]{0.38\linewidth}
            \centering
            \includegraphics[width=0.99\linewidth]{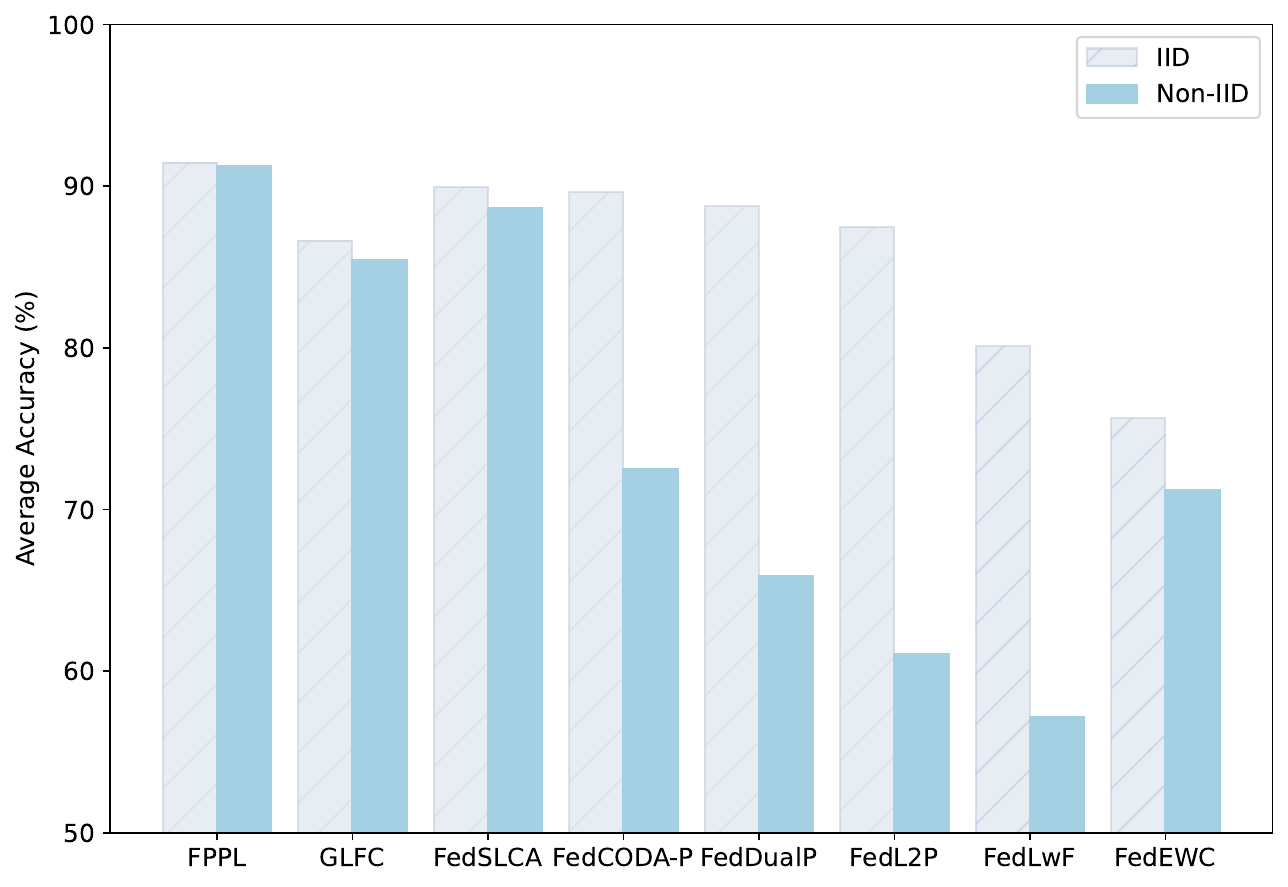}
            \label{fig:cifar100_avgacc}
        \end{minipage}
    }
    \caption{
        Figure \ref{fig:incacc} shows accuracy curves on 20-task ImageNet-R (left) and 20-task CUB-200 (right) during the whole FCL procedure.
        Figure \ref{fig:cifar100_avgacc} presents average accuracy on 10-task CIFAR-100 under IID and non-IID setting.
    }
    \label{fig:overallfig}
\end{figure*}

For ImageNet-R and CUB-200, Table~\ref{tab:overallperformance} presents data across four metrics:
average accuracy $\bar{A}$, last stage accuracy $A_{T}$, average forgetting $\bar{F}$, and communication cost $COMM$;
Figure~\ref{fig:incacc} displays the accuracy $A_t$ at each task stage $t$.

Specifically, in the experiments outlined in Table~\ref{tab:overallperformance} and Figure~\ref{fig:incacc},
non-IID degree was fixed at $\beta=0.5$, and the 200 classes of ImageNet-R and CUB-200 were partitioned into $T=20$, with each task comprising 10 classes.
In Table~\ref{tab:overallperformance}, the performance metrics of FPPL approach optimal levels.
It is notable that FedL2P consistently exhibits lower average forgetting;
however, it is crucial to highlight that its accuracy is noticeably lower.

This is due to its diminished learning capacity, making it ineffective at acquiring knowledge from later tasks.
Consequently, although it demonstrates lower average forgetting for earlier tasks, it also exhibits lower overall global accuracy,
contradicting the primary objective of FCL to mitigate catastrophic forgetting while adapting to new tasks.
In contrast, FPPL achieves both high accuracy and low average forgetting, effectively fulfilling the purpose of FCL.

Additionally, it is observed that methods relying on full fine-tuning,
compared to prompt-based methods, exhibit higher average forgetting $\bar{F}$.
Nevertheless, both FedSLCA and GLFC achieve higher accuracy than other prompt-based methods, with the exception of FPPL.
This discrepancy arises because full fine-tuning on the representation layer enables the model to better adapt to new tasks;
however, continual changes to the representation layer result in catastrophic forgetting for previous tasks.
In contrast, FPPL, by freezing task-specific prompts corresponding to previous tasks and utilizing fused prompt,
significantly alleviates catastrophic forgetting while rapidly adapting to new tasks, resulting in higher accuracy and lower average forgetting.

Figure~\ref{fig:incacc} supports these empirical findings, illustrating that
FedSLCA and GLFC both exhibit higher accuracy in the early stages on 20-task ImageNet-R when there are fewer classes.
This is attributed to the overall fine-tuning of the representation layer,
enhancing learning ability, especially when dealing with challenging instances of the ImageNet-R dataset.
However, this process causes accuracy to gradually decline with an increasing number of learning tasks.
In contrast, FPPL tends to maintain a stable state, consistently achieving high accuracy on 20-task ImageNet-R and CUB-200.

Similar results were obtained on CIFAR-100, as shown in Figure~\ref{fig:cifar100_avgacc}.
The model's average accuracy $\bar{A}$ was examined under IID and non-IID settings with $T=10$, and non-IID degree was set to $\beta=0.5$.
Under IID setting, prompt-based methods performed admirably, comparable to and even surpassing full fine-tuning methods.
However, the non-IID setting caused a significant performance decline compared to IID,
resulting in lower accuracy than full fine-tuning methods, consistent with observations on ImageNet-R and CUB-200.
FPPL maintained consistently high average accuracy $\bar{A}$ across both IID and non-IID settings,
with only a marginal $0.21 \%$ performance decrease.

\subsection{Analysis of Efficiency}

This section presents a comprehensive efficiency analysis of FPPL, comparing it with the representative full fine-tuning method GLFC, as summarized in Table~\ref{tab:efficiency}.
The evaluation covers three dimensions: communication cost, extra storage, and computational burden.

\noindent\textbf{Communication Cost.}
For FPPL, the transmitted parameters encompass the task-specific prompt, coslinear, classifier, and prototypes.
The total transmitted parameters can be formulated as $COMM = D \times (N_{all} + T + \frac{N_{all}}{T} + L_p \times M) + N_{all}$,
where $M$ represents the number of layers in which prompts are inserted.

For 10-task CIFAR-100, with $N_{all} = 100$ and $T = 10$, $COMM = 169,060$.
For 20-task ImageNet-R and CUB-200, with $N_{all}=200$ and $T=20$, $COMM = 253,640$.
In the case of GLFC, it requires transmitting not only the entire updated model
but also the gradient of the encoding network (LeNet in the original paper~\cite{dong2022fcil}),
resulting in a total communication cost $COMM = 87,571,816$.

\noindent\textbf{Extra Storage.}
FPPL requires extra storage for frozen task-specific prompts on the client side,
which can be formulated as $ESTG = D \times L_p \times M \times T$.

For 10-task CIFAR-100, with $T = 10$, final extra storage requirement $ESTG = 768,000$.
For 20-task ImageNet-R and CUB-200, $ESTG = 1,536,000$.
In contrast, GLFC maintains an entire old model for knowledge distillation,
which for the ViT-B/16 model results in $ESTG = 85,952,456$.

We further analyze the extra server-side storage,
where FPPL maintains local prototypes in a prototype pool.
Assuming no missing classes across all clients,
the total extra server-side storage requirement after $T$ tasks is $D \times K \times N_{\text{all}}$.
When $K = 10$, for CIFAR-100 with $N_{\text{all}} = 100$,
the required storage is $768 \times 10 \times 100 = 768,000$ (float).
As CIFAR-100 images have a size of $32 \times 32 \times 3 = 3072$ (int), the storage equals:
$768,000$ floats $\times$ $4$ bytes/float $\div$ $3072$ bytes $= 1000$ images.
Similarly, for ImageNet-R and CUB-200 with $N_{\text{all}} = 200$,
and resized images have a size of $224 \times 224 \times 3 = 150,528$ (int), the extra server-side storage amounts to:
$1,536,000$ floats $\times$ $4$ bytes/float $\div$ $150,528$ bytes $\approx 41$ images.

\begin{table}[htp!]
    \caption{
        \textbf{Client-side efficiency comparison between the full fine-tuning GLFC and prompt-based FPPL on 20-task benchmarks}.
        $RF$ indicates whether the method is rehearsal-free.
    }
    \label{tab:efficiency}
    \centering

    \scalebox{0.99}{
        \begin{tabular}{c||c|c|c|c}
            \hline
            \rule{0pt}{10pt} Method & $RF$              & $COMM$ ($\downarrow$)    & $ESTG$ ($\downarrow$)   & $PARAM$ ($\downarrow$)  \\
            \hline
            GLFC                    &                   & $87.57 \times 10^6$      & $85.95 \times 10^6$     & $85.80 \times 10^6$     \\
            \textbf{FPPL}           & $\bm{\checkmark}$ & $\bm{00.25 \times 10^6}$ & $\bm{1.54 \times 10^6}$ & $\bm{0.09 \times 10^6}$ \\

            \hline
        \end{tabular}
    }

\end{table}

\noindent\textbf{Computational Burden.}
During local training on the client side,
FPPL tunes only the task-specific prompt, coslinear, and classifier.
The tunable parameters excluding classifier can be formulated as $PARAM = D \times (L_p \times M + T)$.

With a maximum setting of $T = 20$,
the tunable parameters excluding classifier is $PARAM = 92,160$,
which is just $0.11 \%$ of the parameters in the fully fine-tuned ViT backbone.
For 10-task CIFAR-100, with $T = 10$, tunable parameters excluding classifier is $PARAM = 84,480$.
Conversely, GLFC typically fine-tunes the entire ViT backbone, resulting in $PARAM = 85,798,656$.
Additionally, compared to rehearsal-based methods like GLFC,
FPPL avoids incorporating data from previous tasks during training on the client side,
further reducing computational overhead.

\subsection{Analysis of Non-IID Robustness}

We conducted experiments on ImageNet-R and CUB-200 with varying non-IID degrees $\beta \in \{ 1, 0.5, 0.1, 0.05, 0.01 \}$ to examine the impact on performance.
Our comparisons include the prompt-based method FedCODA-P, full fine-tuning method FedSLCA and GLFC.

Additionally, we further analyzed FPPL's robustness to non-IID by evaluating FPPL-WOP,
which excludes the prototype pool and relies solely on
disposable local prototype set for classifier debiasing.

\begin{figure}[htp!]
    \centering
    \includegraphics[width=0.99\linewidth]{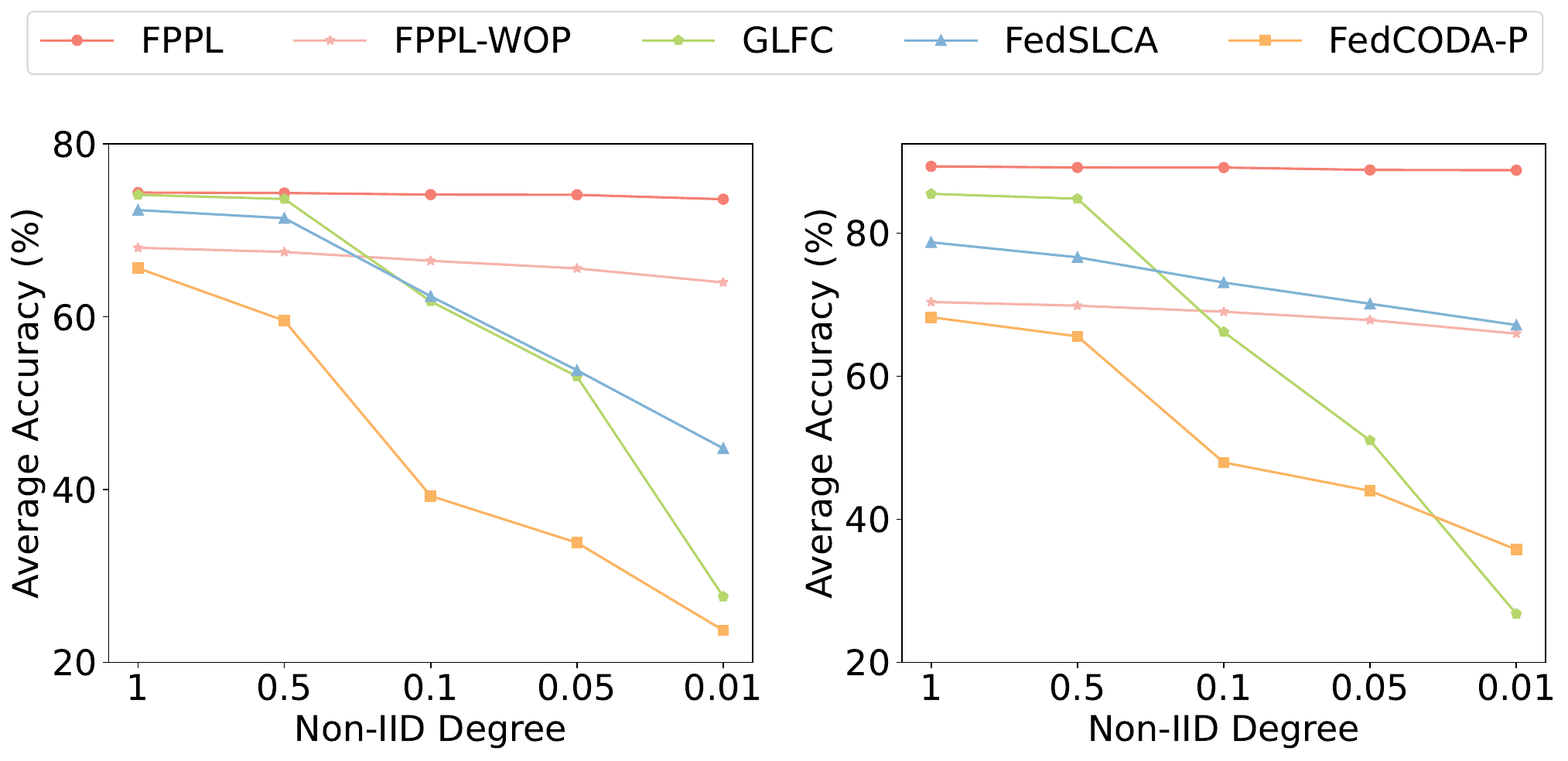}
    \caption{Average accuracy on 20-task ImageNet-R (left) and 20-task CUB-200 (right) under different non-IID degrees.}
    \label{fig:nonIIDness}
\end{figure}

As shown in Figure~\ref{fig:nonIIDness}, FedCODA-P demonstrates sensitivity to non-IID,
with a noticeable performance decrease as $\beta$ diminishes.
Similarly, FedSLCA and GLFC experience a significant performance decline for $\beta \leq 0.1$,
indicating challenges in dealing data heterogeneity under increased non-IID degree.

In contrast, both FPPL and FPPL-WOP demonstrate robustness across various non-IID degrees.
As $\beta$ decreases, FPPL experiences average accuracy drops of $1.03 \%$ on ImageNet-R and $0.59 \%$ on CUB-200,
while FPPL-WOP drops by $5.91 \%$ and $6.27 \%$, respectively.

This robustness can be attributed to the prototype-based debiasing mechanism,
which helps the classifier leverage local data characteristics.
While FPPL-WOP experiences a larger performance drop compared to FPPL due to the absence of a prototype pool designed to mitigate catastrophic forgetting,
it maintains considerable robustness by temporarily using a local prototype set to counteract non-IID issue.

\subsection{Ablation Study}

To further investigate the impact of individual components of FPPL,
we conducted an ablation study on the ImageNet-R dataset, setting $T=20$ and $\beta=0.5$.
Table~\ref{tab:ablations} provides insights into the influence of removing three key components:
unified representation loss, fusion function, and debiasing.

Remarkably, the removal of the debiasing mechanism incurs a significant decline in performance,
emphasizing its pivotal role in rectifying classifier bias.
In FCL scenarios characterized by non-IID and absence of data from previous tasks,
classifiers are susceptible to performance degradation induced by bias.
Our debiasing mechanism, grounded in lightweight prototypes, effectively addressed this challenge.

Additionally, both the unified representation loss and the fusion function contribute to FPPL.
The unified representation loss unifies the representations across clients through contrastive learning between extracted features and global prototypes,
thereby improving accuracy and reducing forgetting.
Similarly, incorporating the fusion function helps to leverage the knowledge encapsulated in all task-specific prompts,
which becomes increasingly crucial as the number of tasks grows.

Moreover, we examine the effect of scaling up the number of clients on performance.
As shown in Table~\ref{tab:client-num}, there is a marginal decline of performance with the increasing number of clients.
Within the same non-IID degree, the expansion of clients exacerbates the data heterogeneity among them, thereby adversely affecting performance.
However, our FPPL demonstrates strong robustness, with a maximum decrease in average accuracy of only $1.82 \%$.

\begin{table}[htp!]
    \caption{
        \textbf{Ablation results on three key components of our FPPL framework}:
        unified representation loss (U.), fusion function (F.), and debiasing mechanism (D.).
        The experiments are conducted on 20-task ImageNet-R.
    }

    \centering
    \label{tab:ablations}

    \scalebox{0.99}{
        \begin{tabular}{c c c|c c c}
            \hline
            \rule{0pt}{10pt} U. & F.                & D.                & $\bar{A}$ ($\uparrow$) & $A_{T}$ ($\uparrow$)  & $\bar{F}$ ($\downarrow$) \\
            \hline
                                &                   &                   & $54.75 \pm 2.60$       & $46.78 \pm 1.78$      & $10.18 \pm 0.91$         \\
            $\bm{\checkmark}$   &                   &                   & $55.89 \pm 2.66$       & $48.78 \pm 1.89$      & $8.74 \pm 0.81$          \\
                                & $\bm{\checkmark}$ &                   & $56.30 \pm 2.28$       & $50.13 \pm 1.67$      & $7.30 \pm 0.88$          \\
            $\bm{\checkmark}$   & $\bm{\checkmark}$ &                   & $59.00 \pm 1.96$       & $52.78 \pm 0.99$      & $5.60 \pm 0.49$          \\
                                &                   & $\bm{\checkmark}$ & $71.82 \pm 0.10$       & $65.63 \pm 0.08$      & $6.45 \pm 0.55$          \\
            $\bm{\checkmark}$   &                   & $\bm{\checkmark}$ & $72.74 \pm 0.13$       & $67.03 \pm 0.16$      & $6.20 \pm 0.34$          \\
                                & $\bm{\checkmark}$ & $\bm{\checkmark}$ & $72.96 \pm 0.45$       & $68.42 \pm 0.40$      & $5.20 \pm 0.16$          \\
            $\bm{\checkmark}$   & $\bm{\checkmark}$ & $\bm{\checkmark}$ & $\bm{74.33 \pm 0.37}$  & $\bm{70.84 \pm 0.15}$ & \bm{$5.07 \pm 0.27$}     \\

            \hline
        \end{tabular}
    }
\end{table}

\begin{table}[htp!]
    \caption{
        \textbf{Performance of FPPL with varying number of clients}.
        The experiments are conducted on 20-task ImageNet-R.
    }

    \centering
    \label{tab:client-num}

    \scalebox{0.99}{
        \begin{tabular}{c|c c c}
            \hline
            \rule{0pt}{10pt} $K$                      & $\bar{A}$ ($\uparrow$) & $A_{T}$ ($\uparrow$)  & $\bar{F}$ ($\downarrow$) \\
            \hline
            $\bm{10 (\textbf{default})}$ & $\bm{74.33 \pm 0.37}$  & $\bm{70.84 \pm 0.15}$ & $\bm{5.07 \pm 0.27$}     \\
            $20$                                      & $73.08 \pm 0.10$       & $69.08 \pm 0.42$      & $6.57 \pm 0.29$          \\
            $30$                                      & $72.98 \pm 0.34$       & $68.40 \pm 0.40$      & $7.74 \pm 0.57$          \\
            $40$                                      & $73.01 \pm 0.50$       & $68.18 \pm 0.22$      & $8.70 \pm 0.55$          \\

            \hline
        \end{tabular}
    }
\end{table}

\begin{table}[htp!]
    \caption{
        \textbf{Performance comparison between Fed-CPrompt and FPPL on 10-task CIFAR-100}.
        Results of Fed-CPrompt are copied from the original paper~\cite{bagwe2023fedcprompt}.
    }
    \label{tab:cprompt}
    \centering

    \scalebox{0.98}{
        \begin{tabular}{c|c c||c c|c}
            \hline
            \multirow{2}{*}{Method} & \multicolumn{2}{c||}{IID} & \multicolumn{2}{c|}{Non-IID} & \multirow{2}{*}{$COMM$ ($\downarrow$)} \\
            \cline{2-5}
            \rule{0pt}{10pt}        & $\bar{A}$ ($\uparrow$)    & $\bar{F}$ ($\downarrow$)
                                    & $\bar{A}$ ($\uparrow$)    & $\bar{F}$ ($\downarrow$)                                              \\
            \hline
            Fed-CPrompt
                                    & $79.43$                   & $\bm{4.75}$
                                    & $65.45$                   & $9.15$
                                    & $00.35\times 10^6$                                                                                \\
            \textbf{FPPL}
                                    & $\bm{89.10}$              & $6.62$
                                    & $\bm{88.92}$              & $\bm{7.39}$
                                    & $\bm{00.25\times 10^6}$                                                                           \\

            \hline
        \end{tabular}
    }

\end{table}

\subsection{Comparison with Fed-CPrompt}

We conducted experiments comparing FPPL with Fed-CPrompt~\cite{bagwe2023fedcprompt},
another rehearsal-free FCL method based on prompts,
while strictly adhering to Fed-CPrompt's experimental settings.
Specifically, compared to our settings in Section~\ref{impdetails},
the learning rate is set to $lr = 1e^{-4}$ and communication rounds per task $R = 40$, with all other settings remaining unchanged.
Table~\ref{tab:cprompt} highlights FPPL's superior performance over Fed-CPrompt in both IID and non-IID settings,
underscoring its robustness and potential as a solution for FCL applications.

\section{Conclusion}
\label{sec:conclusion}

In this work, an efficient and non-IID robust federated continual learning framework,
called {\em{\textbf{F}ederated \textbf{P}rototype-Augmented \textbf{P}rompt \textbf{L}earning (FPPL)}}, is proposed.
FPPL effectively mitigates performance degradation from both non-IID and catastrophic forgetting by leveraging prompts augmented with prototypes.
Specifically, on the client side, a fusion function and a unified representation loss are utilized during local training;
on the server side, we utilize a classifier debiasing mechanism to fully exploit local data characteristics.
Experiments conducted on representative benchmark datasets validate the effectiveness and non-IID robustness of FPPL,
demonstrating notable performance across diverse non-IID degrees with an efficient design.

\bibliographystyle{plainnat}
\bibliography{egbib}

\end{document}